\documentclass[runningheads,a4paper]{llncs}

\usepackage{amssymb}
\usepackage{graphicx}
\usepackage{epstopdf}
\usepackage{url}
\pdfoutput=1
\usepackage{graphicx}
\usepackage{blindtext}
\usepackage{amssymb}
\usepackage{amsmath}
\usepackage{multirow}
\usepackage{subfig}
\usepackage{paralist}
\usepackage{amssymb,color}
\usepackage{algorithmicx}
\usepackage{algorithm}
\usepackage{algpseudocode}
\usepackage{pifont}
\usepackage{hyperref}
\usepackage{url}
\usepackage{float}
\usepackage{cite}

\usepackage{url}
\urldef{\mailsa}\path|{alfred.hofmann, ursula.barth, ingrid.haas, frank.holzwarth,|
\urldef{\mailsb}\path|anna.kramer, leonie.kunz, christine.reiss, nicole.sator,|
\urldef{\mailsc}\path|erika.siebert-cole, peter.strasser, lncs}@springer.com|
\newcommand{\keywords}[1]{\par\addvspace\baselineskip
\noindent\keywordname\enspace\ignorespaces#1}

\begin{document}

\mainmatter  % start of an individual contribution

\title{Orbital Petri Nets: A Novel  Petri Net Approach}

\titlerunning{Lecture Notes in Computer Science: Authors' Instructions}

\author{Mohamed Torky$^{2}$
%\thanks{Please note that the LNCS Editorial assumes that all authors have used
%the western naming convention, with given names preceding surnames. This determines
%the structure of the names in the running heads and the author index.}%
\and $^{1}$Aboul Ella Hassanein $^{1,2}$}
%


\institute{${^2}$Cairo Univesrity, Faculty of Computers and Information, Cairo,Egypt\\Scientific Research Group in Egypt (SRGE)\\
\email{mtorky86@gmail.com} \\
\email{aboitcairo@gmail.com}\\
\url{http://www.egyptscience.net/}}

\maketitle

\begin{abstract}
Petri Nets is very interesting tool for studying and simulating different behaviors of information systems. It can be used in different applications based on the appropriate class of Petri Nets whereas it is classical, colored or timed Petri Nets. In this paper we introduce a new approach of Petri Nets called orbital Petri Nets (OPN) for studying the orbital rotating systems within a specific domain. The study investigated and analyzed OPN with highlighting the problem of space debris collision problem as a case study. The mathematical investigation results of two OPN models proved that space debris collision problem can be prevented based on the new method of firing sequence in OPN. By this study, new smart algorithms can be implemented and simulated by orbital Petri Nets for mitigating the space debris collision problem as a next work.

\end{abstract}

\section{Introduction}

Petri Nets (PN) is mathematical and graphical modeling tools proposed by Car Adam in 1962, which used to study various behaviors of different systems.  They are interesting and popular tools for modeling and studying information processing systems that are characterized as being parallel, distributed, concurrent, asynchronous, nondeterministic, and/or stochastic \cite{Murata 1989}. The classical Petri Nets  is a directed bipartite graph with two node types called places and transitions. Places are represented by circles and transitions by bars.Places may contain zero or more tokens, which drawn as black dots, and the number of tokens may change during the execution of the net. A place $p$ is called an input place of a transition $t$ if there exists a directed arc from $p$ to $t$, and  $p$ is called an output place of $t$ if there exists a directed arc from $t$ to $p$. Always, tokens in the input places represent a specific information in the form preconditions, input data, input signals, or resource needed, whether, tokens in the output places represent post-conditions, output data, output signals, or resource released, . The transition represents a specific function, event, method, computational step, signal processor, processor, algorithms, or clause in logic, whether, Places in Petri Nets works as a repository of input tokens or output tokens. The behavior of system change can be described as set of marking states. Hence, in order to simulate the dynamic behavior of a specific system, a set of marking states in the handled Petri Net model is changed according to the Enabling/Firing  rules of the set of transitions in the net. Figure 1 (a) depicts a classical Petri net model consist of one enabled transition $t$ has two input places $P1, P2$ and three output places $P3, P4$, and $P5$ . the two input places ($P1$, and $P2$) are being marked with two and four tokens respectively. Firing $t$ in Figure 1 (b) consumes one token from $P1$ and two tokens form $P2$, and then produces three, two, and one tokens in $P3, P4$, and $P5$ respectively according to the input and output arc-weights.

%=========== Fig 1 ================
  \begin{figure*}[!t]
\centering
\fcolorbox{black}{white}{\includegraphics[width=1\linewidth]{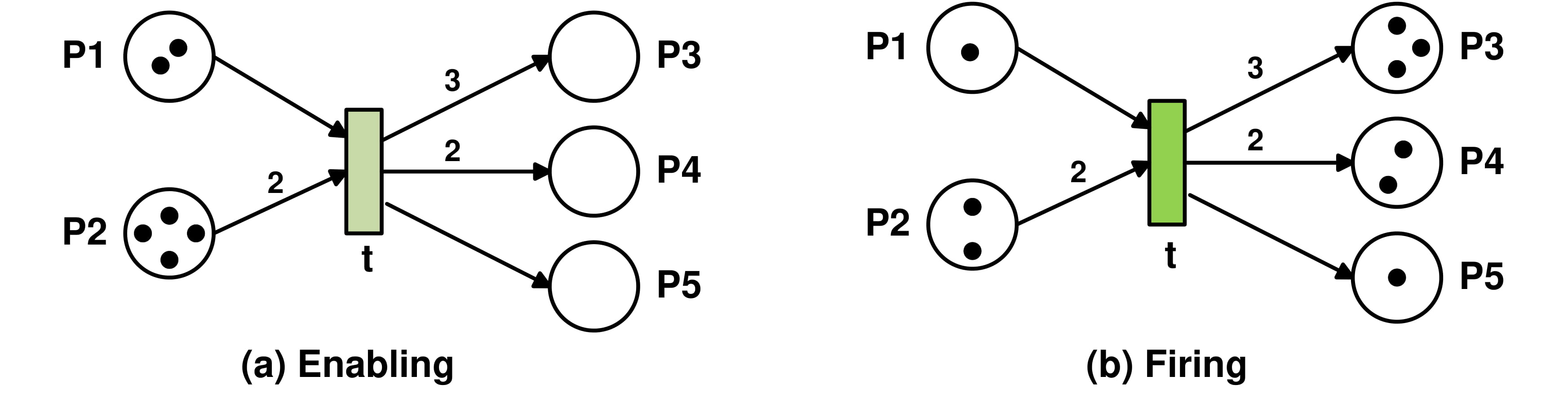}}
\caption{Classical Petri Net: (a) Enabled transition $t$  (b) Firing transition $t$.}
\label{fig:1}       % Give a unique label
\end{figure*}

%==============================

Adding more features and properties to tokens, places, and transitions result in more classes of high level Petri Nets such as Colored Petri Nets \cite{Jensev 2013} , Timing Petri Nets \cite{Van 1996}. In Colored Petri Nets (CPN), tokens is represented  with an attached data value called \emph{token-color}, and Each place represent a \emph{color set}, thereby, for given place $p$ all tokens in $p$ must belong to the color set of $p$.Enabling of transitions in CPN depends on the Binding concept.  Binding operation allows the corresponding transition to fire according to what called  \emph{arc-inspectors}, which define input/output tokens for transitions in the form of expression of variables. For example, Figure 2 depicts the enabling-firing process in CPN. Figure 2 (a) demonstrates that the transition $t1$ has two enabled bindings according to two variables $x$, and $y$. variable $x$ can be substituted with an integer value from $P1$ which may be '8' or '3', and variable $y$ can be substituted with an integer value from $P2$ which has only '6' values. Transition $t1$ is associated with a guard expression $x>y$ that returns Boolean value for allowing firing process. Firing $t1$ as Figure 2 (b) produces a new token in $P3$ which  marked with the value 22 according to an arc-inspector $2x+y$.

In Timing Petri Nets (TPN), time feature  is associated with tokens, and transitions determine delays of tokens. Each token has a time-stamp which models the time the token becomes available for consumption. The time-stamp of a produced token is equal to the \emph{firing time} plus the \emph{firing delay} of the corresponding transition. Figure 3 (a) shows a Petri net which models two identical parallel machines for processing thee jobs , and Figure 3(b) show the firing results of both transitions \emph{start 1}, and \emph{start 2}. by the same firing way,  the firing results of the both further transitions \emph{finish 1}, and \emph{finish 2} will produce 3 tokens in place \emph{out} indicating that the three jobs have been processed successfully. The time-stamps of these tokens are equal to the corresponding completion time of each one which are . 5, 6 and 8 respectively.

%=========== Fig 2 ================
  \begin{figure}[H]
\centering
\fcolorbox{black}{white}{\includegraphics[width=1\linewidth]{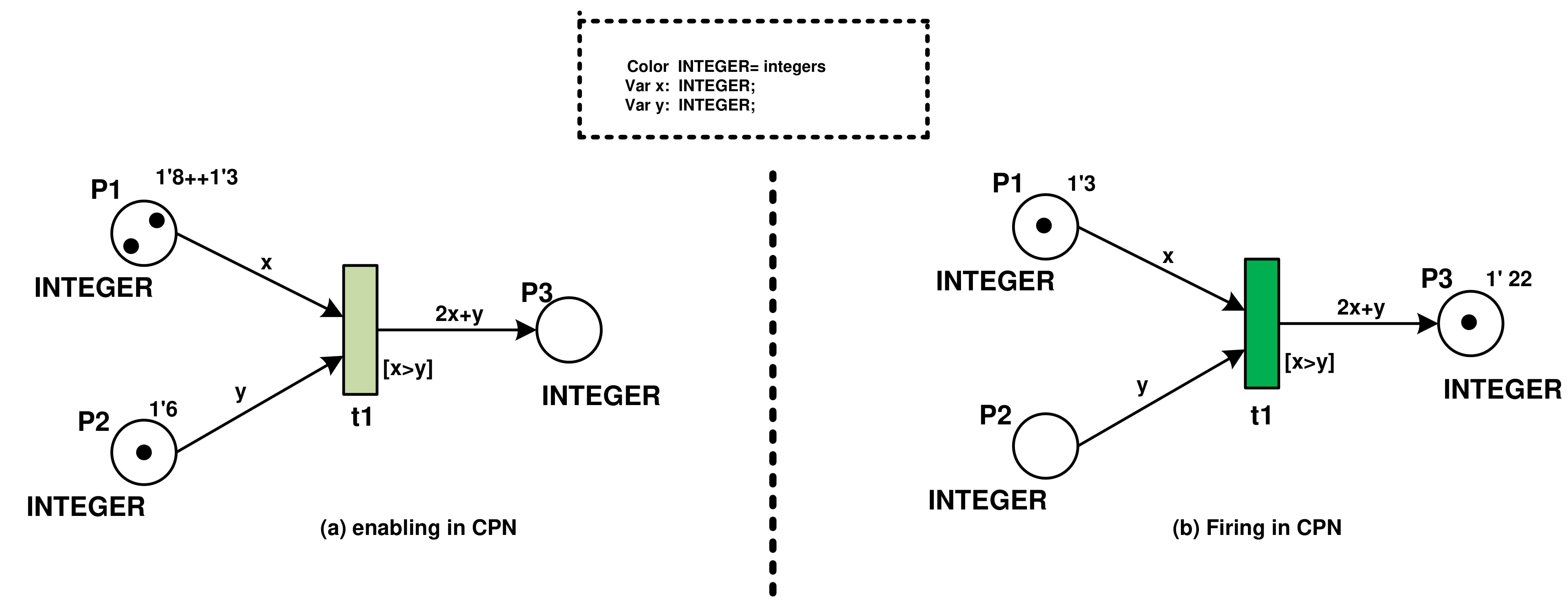}}
\caption{Colored Petri Net (CPN): (a) Enabling transition $t1$  (b) Firing transition $t1$.}
\label{fig:1} % Give a unique label
\end{figure}

%==============================

%=========== Fig 3 ================
  \begin{figure}[H]
\centering
\fcolorbox{black}{white}{\includegraphics[width=1\linewidth]{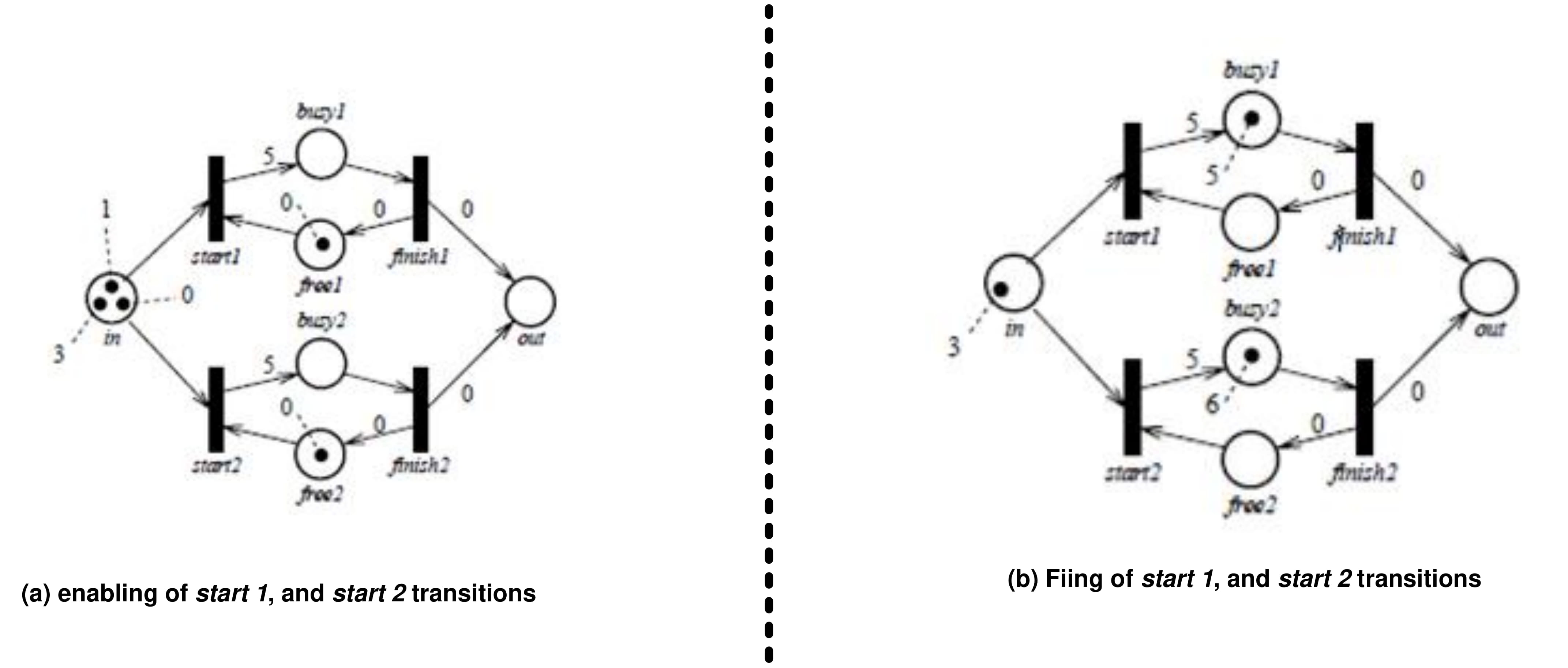}}
\caption{Timing Petri Net (CPN): (a) Enabling transitions $stat 1, and stat 2$  (b) Firing transitions $stat 1, and start 2$.}
\label{fig:1} % Give a unique label
\end{figure}

%==============================

Although the existential of variety classes of Petri Nets for studying systems that are parallel, distributed, concurrent, asynchronous, nondeterministic, and/or stochastic, there is no a Petri Net appoach for studying the behavior of rotating systems within a specific domain. In this paper, we propose a novel approach of Petri Nets called Orbital Petri Nets (OPN)for studding and simulating orbital rotating systems, such as orbital movement of  satellites in space science applications or studying the orbital movements of electrons  in the natural sciences. The feasibility of OPN has been analyzed and verified on two OPN models for simulating preventing two satellite collision problem as well as preventing a satellite and debris collision problem. The rest of this paper can be arranged as section 2 presents the recent applications of Petri Nets in the literature. Section 3 presents the proposed Orbital Petri Net approach, section 4 presents the mathematical representation of OPN.  Section 5 presents a case study  of applying OPN in space collision problem. Section 5 formulates the conclusion of this work.

\section{Related Work}

The recent literature on Petri Nets has highlighted several applications in different fields. Fuzzy Petri nets (FPNs)\cite{Liu 2017}  has been introduced for modeling mechanisms for knowledge representation and reasoning algorithms of rule-based expert systems. An adaptive Petri net (APN)\cite{Ding 2016}  is introduced to model a self-adaptive software system with learning Petri Nets. APN is an extension of hybrid Petri nets by embedding a neural network algorithm into them at some special transitions. Controllable Siphon Basis of Petri Nets \cite{Liu H 2015}  has been used to prevent deadlock for flexible manufacturing systems. Moreover,  a novel Petri Net model is proposed \cite{Ding Z 2015}  for analyzing the performance of concurrent system based on a polynomial algorithm and ordinary differential equations. Jianhong Ye et al \cite{Ye J 2015}   introduced a decentralized supervision policy for a Petri net through collaboration between a coordinator and subnet controllers as a discrete event system (DES). Timed Colored Petri Nets \cite{Bauwa 2015}  approach has been proposed for addressing the deadlock (DL)-free scheduling problem of flexible manufacturing systems (FMS). Zhou etal \cite{Zhou Y 2000}  proposed a fuzzy-timing Petri nets for modeling and analyzing the performance of networked virtual environments (net-VEs). This paper proposes to apply a Petri net formal modeling technique on a net-VE-NICE (narrative immersive constructionist/collaborative environment), predict performance of  the net-VE  based on simulation, as well as  improving the net-VE performance.  Analyzing the compatibility of Web services under temporal constraints \cite{Du Y 2014}  is another promising approach that utilized the timing Petri Net in a holistic manner and in modular way for modeling and analyzing the problem of message mismatches between services in a composition. Yi-Sheng Huang et al \cite{Huang 2014}   proposed a new approach to design and analyze an urban traffic network control system by using the Synchronous Timing Petri Nets (STPNs). The advantage of the proposed approach reflected a clear presentation of the behaviors of traffic lights in terms of the conditions and events that cause phase alternations. Moreover, the size of the urban traffic network control system can be easily extended with the proposed modular technique. Modeling and validating an e-commerce system is another trend of Petri Net applications \cite{Yu W 2014} . In this study, the authors introduced a Petri Net model for constructing an e-commerce business process called an E-commerce Business Process Net, which integrates both data and control flows for modeling and analyzing the communication events in e commerce systems. Modeling and simulating Reaction–diffusion systems in systems biology is another application of Petri Nets \cite{Liu F 2014} . The authors investigated a colored Petri net framework integrating deterministic, stochastic and hybrid modeling formalisms and corresponding simulation algorithms for simulating the reaction–diffusion processes that may be closely coupled with signaling pathways, metabolic reactions and/or gene expression.  Baldin  et al \cite{Bonilla 2014}  utilized a colored Petri Nets for modeling the concurrent an distributed nature for designing a Coordinated human-wearable robot control on the shoulder. The CPN framework is then extended to a type of hybrid control system by imbedding local dynamic controllers in the Transition nodes of the CPN model. Each local dynamic controller collects sensor signals relevant to the target transition and makes a predictive control decision. This allows the robot on the shoulder to take a proactive and preemptive action as well as to confirm a successful Transition. V. M. Teslyuk etal \cite{Teslyuk 2013}   utilized the colored Petri Nets for developing a smart house system model, which  enabled exploring dynamics of the whole system as well as internal interaction of its main structural and functional subsystems at the system level design, have been developed. On anothe study, a Colred Petri Net model is proposed for modeling a Secure interoperation design in multi-domains environments \cite{Kirchner H 2013} . The study addressed several types of potential conflicts and consistency properties with a systematic and rigorous approach: graph theory, network flow technology and colored Petri nets are applied for specifying and verifying a secure interoperation design.

\section{Orbital Petri Nets Approach}

One of the most well-known tools for modeling and simulating system behaviours is Petri Nets. Although the variety of Peti Net classes, there is no Petri Net approach can be utilized for studying orbital rotating system. In response to this need, we introduce new Petri Net approach called Orbital Petri Nets (OPN) for studying and simulating the behaviour of rotating systems in cyclic orbits such as satellite and aircraft movement in different orbits around the earth as depicted in Figure 4. In the graphical representation of orbital Petri Nets, places are drawn as directed circle for representing directed cyclic orbits in the handled domain. Tokens  as  colored dots for representing the rotating objects in the orbits (e.g. satellites or electrons, etc). Transitions as bars or boxes for representing processing function (or method) on the rotating tokens on the cyclic places. Arcs are labeled with their weights expressions that define tokens transfer from input places to the output places. The arc-weight may be omitted if all arcs have the default weight;   The marking in OPN assigns number of tokens in each cyclic place (i.e. number of rotating tokens in the orbit). A marking $ M_i$ assigns to a cyclic place $p$ a positive integer k-tokens. A marking $ M_i$ is represented as an m-vector, where $m$ is the total number of places in the handled system. Each transition $t$ has input and output cyclic places. A transition $t$ is said to be enabled if there is any token in any input  place can be called on the input arc weight. Firing a transition $t$ moves the rotated tokens from input places to the output places according to input/output arc-weights callings with respect to the boolean expression associated with $t$.

%=========== Fig 3 ================
  \begin{figure}[!t]
\centering
\fcolorbox{black}{white}{\includegraphics[width=1\linewidth]{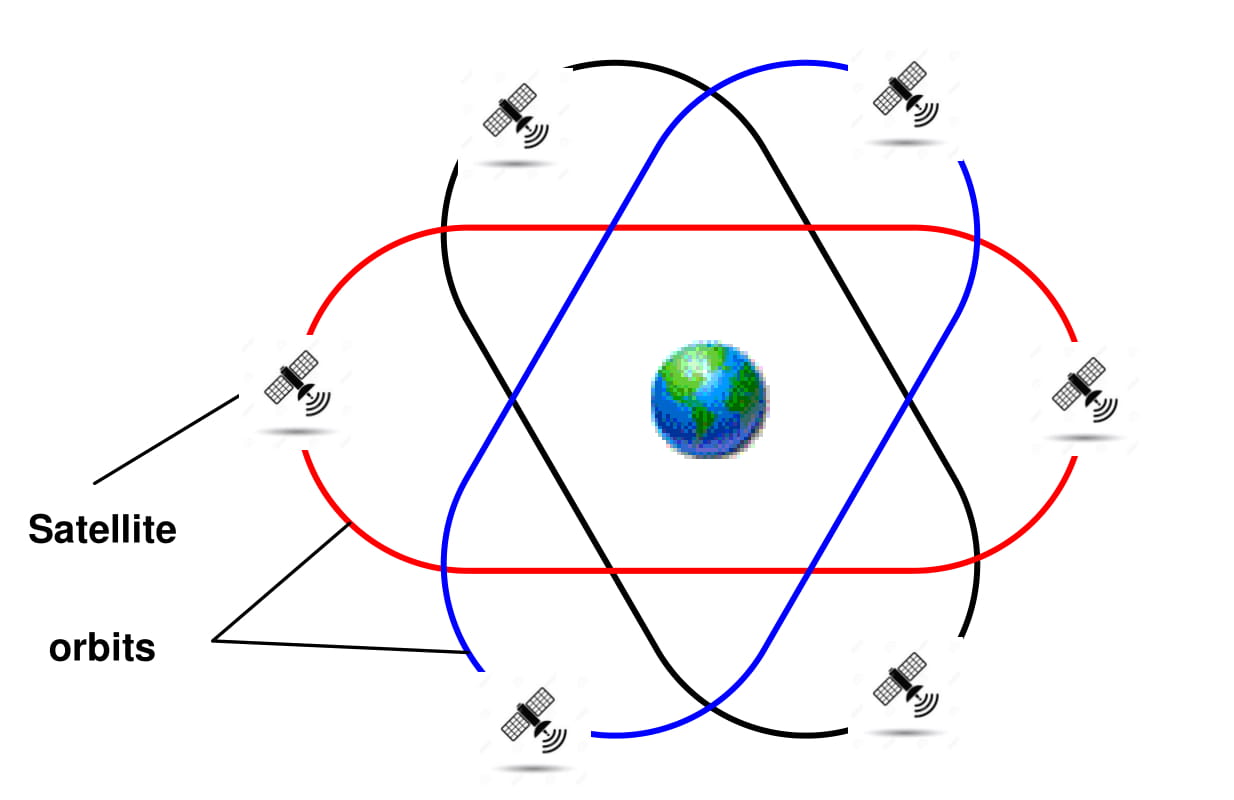}}
\caption{orbital movements of satellites around the Earth.}
\label{fig:1} % Give a unique label
\end{figure}

%==============================

 The order of a given OPN model is specified by the number of cyclic places in the model (i.e. number of orbits in the given domain). It is important here to clear that enabling/firing rule in OPNs is different from the other classes of Petri Nets. The known idea about enabling/firing rule of transition $t$ is consuming tokens from input places and producing other tokens in the output places according to the input and output arc-weights, where  $M(p)\geq w(p,t)$.(i.e. number of tokens in a specific input place is greater than or equal to arc-weight from place $p$ to transition $t$ ). On the other hand, In OPN enabling/firing rule of transition $t$ moves the tokens in any input place to the output place according to token calling in the input/output arc-weights expressions. An  interesting advantage of OPN is that OPN may import some features of other known classes of Petri Nets such as Timing, Coloring, or inhibitor arcs with respect to the behaviour of the handled system.\\
 The Formal definition of Orbital Petri Nets (OPN) of order $N$ can be formulated as follows:\\

 OPN=$(P^{+/-},T,A,\sum,W,G,M_0 )$   where\\
 \begin{itemize}
   \item $P^{+/-}=\{ p_1^{+/-}, p_2^{+/-},  p_3^{+/-},...\}$ is a finite set of directed cyclic places. The signs  +/-  represent the rotation direction of tokens in places(i.e. clockwise '+' or anticlockwise '-').
   \item $T=\{t_1,  t_2,t_3,…..\}$   is a finite set of transitions.
   \item $A\subseteq(P\times T)\cup(T\times P)$  is a set of input/output arcs
   \item $\sum$  is the set of all types (or set of colores $\{c_1,c_2,..c_m\}$) of rotated tokens  in the set of
         directed places
   \item $W:F \rightarrow \{e_1,e_2,e_3,..\}$  is a weight function,where $e_{j }$ is weight expression that
         defines the method of tokens movement with respect to the token calling in each weight expression $e_{j }$
   \item $G$ is a guard function that  It maps each transition $t\in T$ to a boolean guard expression $g$.
   \item $M_0:P_j^{+/-}\rightarrow \{(ac_1 \cup bc_2,....\cup dc_m )\}$   is the initial marking state of number token types(or colors) in all places.
 \end{itemize}

 \textbf{Example 1:} suppose we have two objects are rotating in different orbits within a specific domain as in Fig 5 (a), and we need to exchange the orbits of each one infinitely. In other words if the object $x$ is rotating in an orbit $A$, and object $y$ is rotating in an orbit $B$, how to swap the orbits of  $x$,and $y$ infinitely using an orbital Petri Net Model? Fig 5(b) depicts the enabling state of transition $t1$, and Fig 5(c) depicts the first round of swapping tokens $x$ and $y$ after firing  $t1$.

 \textbf{Example 2:} suppose we have four objects are rotating in different directed orbits as depicted in Fig. 6(a). what is the OPN model that can represent classifying  rotating objects  into  two classes for preventing the collisions between rotating objects. Fig. 6(b) depicts the constructed OPN model of order 6, and Fig 6(c) depicts the firing process according to input/output arc expressions calling.

 %=========== Fig 5 ================
  \begin{figure}[H]
\centering
\fcolorbox{black}{white}{\includegraphics[width=1\linewidth]{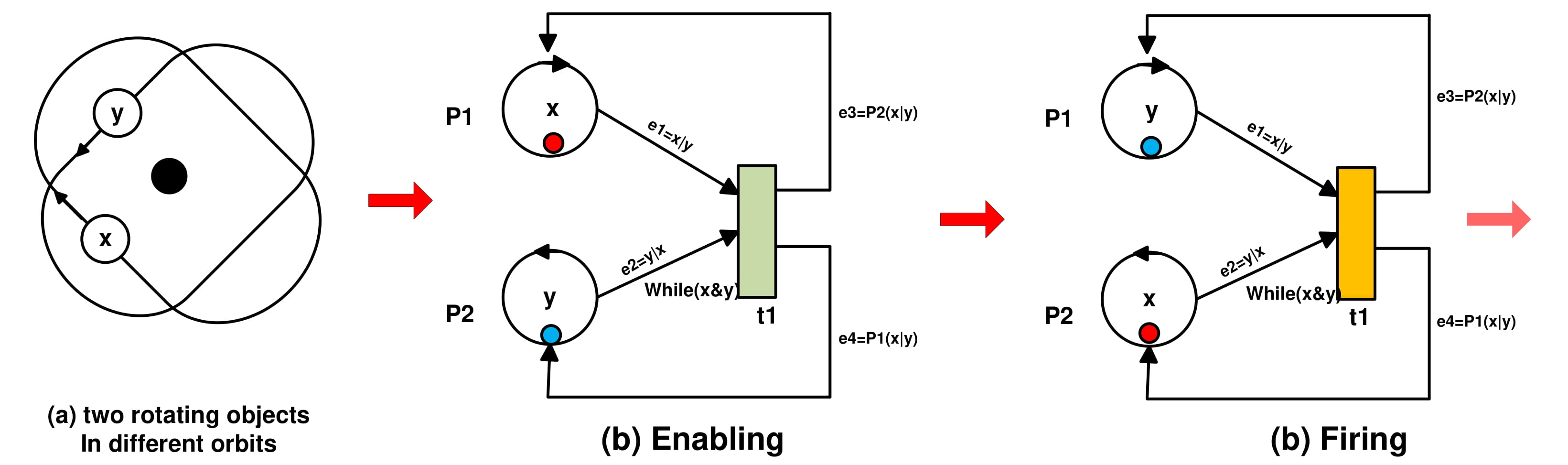}}
\caption{infinite Orbital Petri Net of order 2 (a)two rotating objects in different orbits ,(b) enabling of $t1$, (c) Firing  $t1$}
\label{fig:1} % Give a unique label
\end{figure}

%==============================

%=========== Fig 6 ================
  \begin{figure}[H]
\centering
\fcolorbox{black}{white}{\includegraphics[width=1\linewidth]{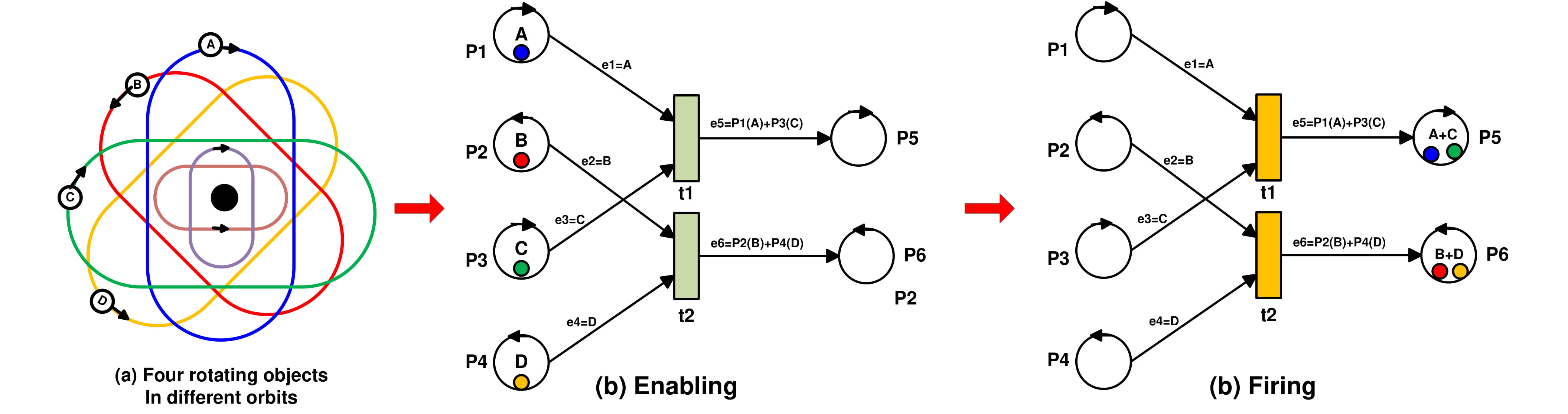}}
\caption{Orbital Petri Net of order 6 (a) four rotating objects in different orbits ,(b) enabling of $t1$ and $t2$, (c)Firing  $t1$ and $t2$}
\label{fig:1} % Give a unique label
\end{figure}

%==============================

\section{Mathematical Representation of Orbital Petri Nets  }

The dynamic behavior of OPN models can be represented by some algebraic equations. In this spirit, we can use \emph{Incidence Matrix} and \emph{State Equations} to govern the dynamic behavior of orbital systems that can be modeled by OPN.\\

\emph{Incidence Matrix:} For orbital Petri nets of order $N$ with $n$ transition and $m$ directed places. The Incidence Matrix $A=[a_{ij}]$ is $n\times m$ matrix of values and its typical entry is given by:

%====== EQ 1==============
\begin{equation}\label{eq 1}
  a_{ij}=a_{ij}^{+}-a_{ij}^{-}
\end{equation}
%=====================

Where, $a_{ij}^{+}$ is the weight value from the transition $i$ to its output place $j$ and $a_{ij}^{-}$ is the weight value of the arc to transition $i$ from its input place $j$ .  According to the firing rule of Orbital Petri Nets, $a_{ij}^{-}$ is the number of tokens moved from the input place of transition $i$,  $a_{ij}^{+}$ is the number of token added to the output place of transition $i$,  and $a_{ij}$ is the change of tokens in place $j$ after firing the transition $i$. The transition $i$ is enabled  at marking state $M$ if there exist place $j$ defines the following:

%====== EQ 2==============
\begin{equation}\label{eq 2}
  M(j)\geq 1,~~ and~~ M(j)\equiv w(j\rightarrow i)
\end{equation}
%=====================

where, $M(j)$ is the marking (i.e. number of tokens) of place $j$, and$ w(j\rightarrow i)$ is the arc weight from place $j$ to transition $i$. \\

\emph{State Equation:} In orbital Petri Nets, a marking state $M_k$ is an $m\times 1$ column vector. The $j^{th}$  entry of  $M_k$ represents the number of tokens in place $j$ immediately after the $k^{th}$ firing in some firing sequence. The control vector $u_k$ is $n\times1$ column vector which represent number of firing times of transition $i$ in the $k^{th}$ firing. The change from marking state $M_{k-1}$ to $M_k$ as a result of firing transition $i$  can be formulated as in equation 3:

%====== EQ 3==============
\begin{equation}\label{eq 3}
  M_{k}=M_{k-1}+A\times u_{k}
\end{equation}
%=====================

\emph{Necessary Reachability Condition:} suppose that the destination marking state $M_d$ is reachable from the initial marking state $M_0$ through the firing sequence $\{u_1,u_2,…u_d\}$, the state equation 3 can be reformulated as in equation 4:

%====== EQ 4==============
\begin{equation}\label{eq 4}
  M_{d}=M_{0}+A\times \Sigma_{k=1}^{d} u_{k}
\end{equation}
%=====================

Which can be rewritten as:

%====== EQ 5==============
\begin{equation}\label{eq 5}
  \Delta M=A\times X
\end{equation}
%=====================
Where, $\Delta M=M_d-M_0$ and $X= \Sigma_{k=1}^{d} u_{k}$.\\

\textbf{Example 3:} for the orbital Petri Net model shown in Fig. 6, the initial marking  $M_0=(A,B,C,D,0,0)^T$. We can solve state equation to prove that  $M_1$   is reachable from $M_0$ after firing $t1$ and $t2$ one time as follows:

%====== EQ 6==============
\begin{equation}\label{eq 6}
 M_1=\left(
  \begin{array}{c}
    P_1 \\
    P_2 \\
    P_3 \\
    P_4 \\
    P_5 \\
    P_6\\
  \end{array}
\right) = \left(
  \begin{array}{c}
    A \\
    B \\
    C \\
    D \\
    0 \\
    0\\
  \end{array}
\right) +\left(
            \begin{array}{cc}
              -A~&~0 \\
               0~&~-B \\
              -C~&~0 \\
               0~&~-D \\
              A+C~&~0 \\
               0~&~B+D \\
            \end{array}
          \right) \times \left(
                            \begin{array}{c}
                              1 \\
                              1 \\
                            \end{array}
                          \right)
\end{equation}
%=====================

Hence,

%============ eq 7========
\begin{equation}\label{eq 7}
 M_1= \left(
  \begin{array}{c}
    P_1 \\
    P_2 \\
    P_3 \\
    P_4 \\
    P_5 \\
    P_6\\
  \end{array}
\right) = \left(
  \begin{array}{c}
    A \\
    B \\
    C \\
    D \\
    0 \\
    0\\
  \end{array}
\right) +\left(
  \begin{array}{c}
    -A \\
    -B \\
    -C \\
    -D \\
    A+C \\
    B+D \\
  \end{array}
\right)=\left(
          \begin{array}{c}
            0 \\
            0 \\
            0 \\
            0 \\
            A+C \\
            B+D \\
          \end{array}
        \right)
\end{equation}

So it is easy to prove that:\\
\begin{equation}\label{eq 8}
  \Delta M=M_1-M_0 =A \times X=\left(
  \begin{array}{c}
    -A \\
    -B \\
    -C \\
    -D \\
    A+C \\
    B+D \\
  \end{array}
\right)
\end{equation}

Hence, this prove that $M_1$ is reachable state from $M_0$

\section{Case Study: Space-Debris Collision Problem}

Orbital space debris has received widespread attention from space-related institutions due to the catastrophic risks on spacecraft and satellites movements. Most of these debris run at 8 km /s (approximately 28,800 km / h). And so quickly these debris can penetrate the structure of spacecraft and pose a threat to the lives of astronauts. It is estimated that a pea-sized body travels so fast has a collision force equal to a body weight of 181 kilograms (400 pounds) runs at 100 km / h (60 m/h). The previous studies demonstrated that there are approximately 5.5 million kilograms of human waste by the earth's orbits. NASA expects that there will be an increasing  in the number of space objects in low Earth orbit by 75 \% over the next 200 years if space debris reduction measures are not followed \cite{liou JC 2008}. The problem of debris disposal is a major and open challenge for researchers and international space stations. Some studies tried to mitigate this problem as in \cite{Nishida 2009}  and \cite{Ker  2008} , however, the recent solutions in the literature didn't introduce an effective way to prevent the collision between space debris and spacecrafts. Also, there is no effective mechanism for space debris disposal for cleaning satellite orbits in the space. In response to this pressing problem, we can utilize the proposed Orbital Petri Nets for modeling and simulating new models for solving space Debris collision problem. The space collision problem can emerge through two scenarios, Spacecraft-Spacecraft collision,and Spacecraft-debris collision. Two orbital petri net models have been introduced for simulating the solutions for these two cases.

\subsection{Case1: Spacecraft-Spacecraft Collision}

The case of  collision two spacecrafts  has been occurred since February 2009 when two satellites have been collided, one American operated by Iridium Satellite (LCC) and the other by Russian for espionage purposes. This collision resulted in 500 to 600 debris in different shapes. After this incident, NASA measured the probability of a serious accident to launch the space shuttle and found that there is a possibility of a serious accident for every 318 shuttle launches \cite{Irene Klotz  2009}. This case of Spacecraft-Spacecraft collision prevention can be modeled by orbital Petri Net as in Fig.7. Fig.7(a) shows the possibility of collision of two satellites in different two intersected orbits. Fig.7(b) depicts the orbital Petri Net model for simulating the collision avoidance between the two satellites. Firing transition $t1$ depends on achieving four conditions , (1)  $M(P1)\geq 1$ and $ M(P1)\equiv w(p1\rightarrow t1)$. (2)  $M(P2)\geq 1$ and $ M(P2)\equiv w(P2\rightarrow t1)$ . (3) the probability of collision is more than 0, (4) the firing time is $T=T1$. Firing results of $t1$ is depicted in Fig. 8(a), such that the two satellite exchange the orbits of each other, hence, token $y$ moved to $P1$, and token $x$ moved to $P2$. Firing transition $t2$ is a consecutive action to firing $t1$, such that, they execute together a smart maneuver for swapping each other without making a collision. Firing $t2$ return each satellite to its correct orbit, hence the token $x$ is returned back to $P1$, and token $y$ is returned back to $P2$ as in Fig. 8 (b). The firing conditions of t2 are; (1)$M(P1)\geq 1$ and $ M(P1)\equiv w(p1\rightarrow t2)$. (2) $M(P2)\geq 1$ and $ M(P2)\equiv w(P2\rightarrow t2)$. (3) the firing time is $T2$, where $T2-T1\leq \varepsilon (T)$

%=========== Fig 7 ================
  \begin{figure}[H]
\centering
\fcolorbox{black}{white}{\includegraphics[width=1\linewidth]{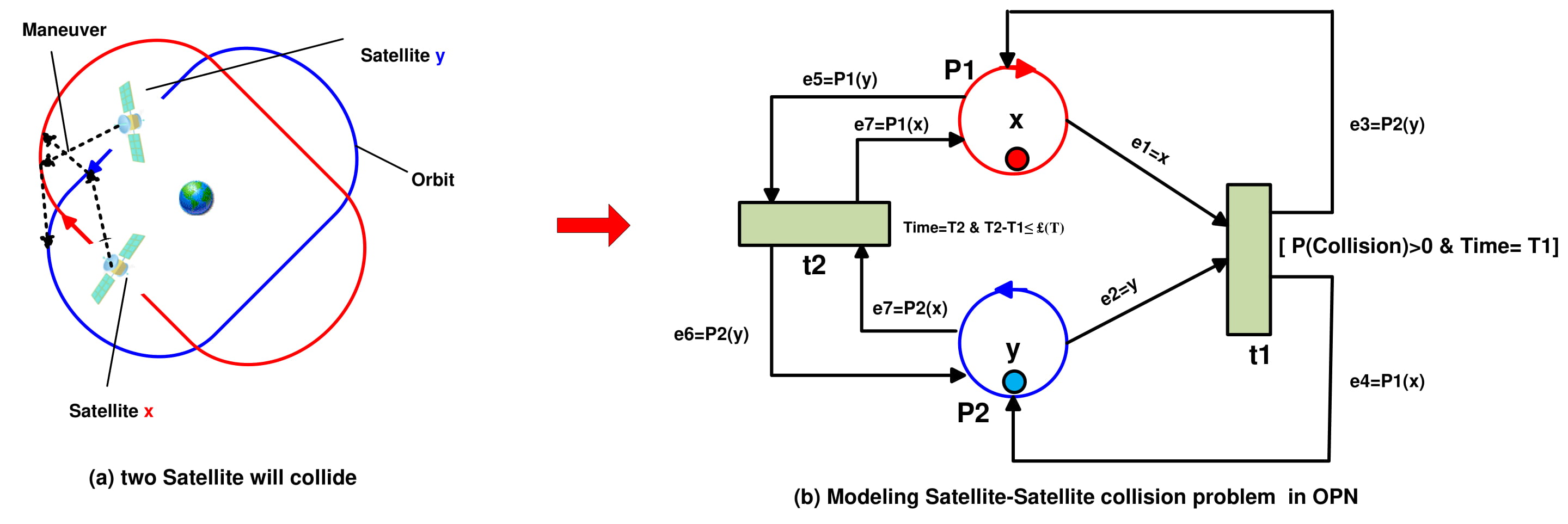}}
\caption{Modeling spacecraft-spacecraft collision problem in Orbital Petri Net}
\label{fig:1} % Give a unique label
\end{figure}

%==============================

%=========== Fig 8 ================
  \begin{figure}[H]
\centering
\fcolorbox{black}{white}{\includegraphics[width=1\linewidth]{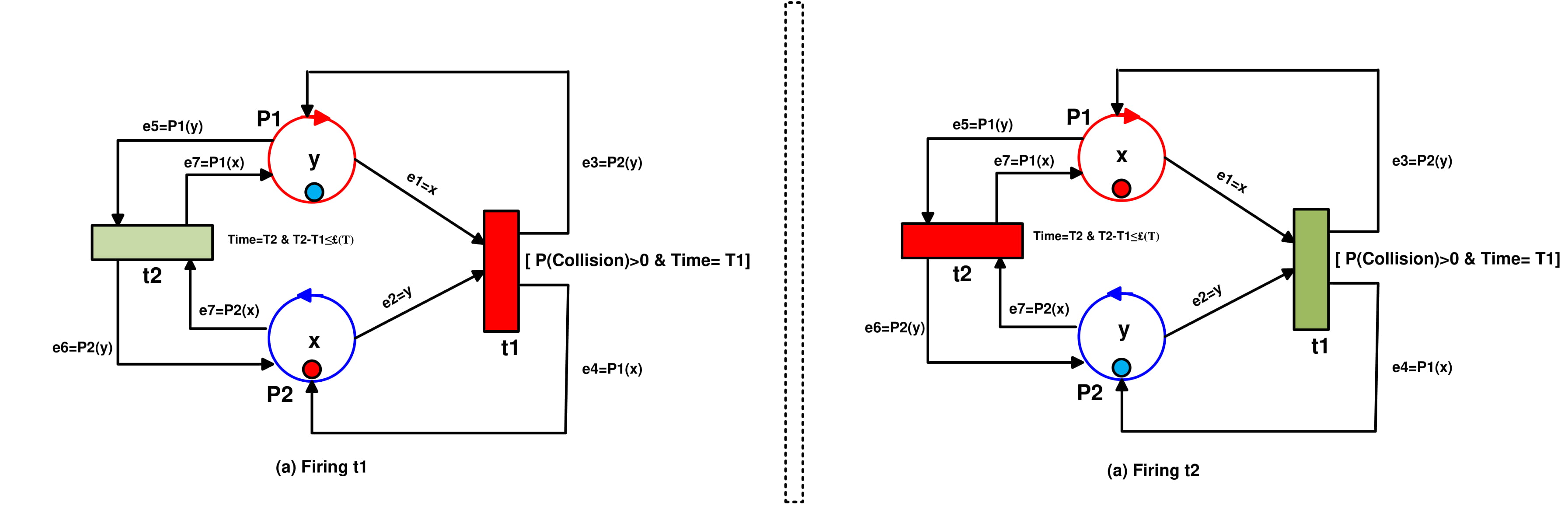}}
\caption{spacecraft-spacecraft collision prevention: (a) Firing $t1$ and (b) firing $t2$ }
\label{fig:1} % Give a unique label
\end{figure}

%==============================

Mathematically, we can represent the collision prevention by proving the Necessary Reachability Condition , thereby, we should prove:

%====== EQ 9==============
\begin{equation}\label{eq 4}
  M_{2}=M_{0}+A\times \Sigma_{k=1}^{2} u_{k}
\end{equation}
%====================

Firing transition $t1$, will produce the marking state  $M_1$ , which can be formulated as:

%====== EQ 10==============
\begin{equation}\label{eq 4}
  M_1=\left(
        \begin{array}{c}
          P_1 \\
          P_2 \\
        \end{array}
      \right)=\left(
                \begin{array}{c}
                  x \\
                  y \\
                \end{array}
              \right)+\left(
                        \begin{array}{cc}
                          y-x &~~ x-y \\
                          x-y &~~ y-x \\
                        \end{array}
                      \right)\times \left(
                                      \begin{array}{c}
                                        1 \\
                                        0 \\
                                      \end{array}
                                    \right)
\end{equation}
%====================

SO,

%====== EQ 11==============
\begin{equation}\label{eq 4}
  M_1=\left(
        \begin{array}{c}
          P_1 \\
          P_2 \\
        \end{array}
      \right)=\left(
                \begin{array}{c}
                  x \\
                  y \\
                \end{array}
              \right)+\left(
                        \begin{array}{c}
                          y-x  \\
                          x-y  \\
                        \end{array}
                      \right)= \left(
                                 \begin{array}{c}
                                   y \\
                                   x \\
                                 \end{array}
                               \right)
\end{equation}
%====================

$M_1$ in equation 11 represents tokens exchange in $P1$ and $P2$ (i.e. satellite swapping in the orbits of each other).\\
Firing transition $t2$, will produce the marking state  $M_2$ , which can be formulated as:

%====== EQ 12==============
\begin{equation}\label{eq 4}
  M_2=\left(
        \begin{array}{c}
          P_1 \\
          P_2 \\
        \end{array}
      \right)=\left(
                \begin{array}{c}
                  y \\
                  x \\
                \end{array}
              \right)+\left(
                        \begin{array}{cc}
                          y-x &~~ x-y \\
                          x-y &~~ y-x \\
                        \end{array}
                      \right)\times \left(
                                      \begin{array}{c}
                                        0 \\
                                        1 \\
                                      \end{array}
                                    \right)
\end{equation}
%====================

Hence,

%====== EQ 13==============
\begin{equation}\label{eq 4}
  M_2=\left(
        \begin{array}{c}
          P_1 \\
          P_2 \\
        \end{array}
      \right)=\left(
                \begin{array}{c}
                  y \\
                  x \\
                \end{array}
              \right)+\left(
                        \begin{array}{c}
                          x-y  \\
                          y-x  \\
                        \end{array}
                      \right)= \left(
                                 \begin{array}{c}
                                   x \\
                                   y \\
                                 \end{array}
                               \right)
\end{equation}
%====================

it is easy to verify and prove the necessary reachability condition in  equation 9 as:

%====== EQ 14==============
\begin{equation}\label{eq 4}
 M_2= \left(
     \begin{array}{c}
             x \\
             y \\
         \end{array}
              \right)+\left(
                        \begin{array}{cc}
                          y-x &~~ x-y  \\
                          x-y &~~ y-x \\
                        \end{array}
                      \right)\times \left(
                                 \begin{array}{c}
                                   1 \\
                                   1 \\
                                 \end{array}
                               \right)=\left(
                                         \begin{array}{c}
                                           x \\
                                           y \\
                                         \end{array}
                                       \right)
\end{equation}
%====================

hence, $M_2$ is reachable state from $M_0$,which represent returning back each satellite to its valid orbit after applying a smart maneuver by the consecutive firing of $t1$ and $t2$  for avoiding the collisions between each other.

\subsection{Case 2: Spacecraft-Debris Collision}

The case of collision of satellite with debris has been occurred since June 2007 when the Republic of China tested anti-satellite missiles, fired a solid-fuel rocket from the Zigang base to hit a Chinese satellite. This collision resulted in 2,300 to 2,500 space particles, which considered the biggest event in the formation of space debris around the earth. This amount of resulted debris has changed the orbit of  Terra environmental spacecraft to avoid potential collisions with the Chinese space debris, the first time that NASA has had to change the orbit of an aircraft or a satellite \cite{Burger 2018}. Moreover, on March, 2009, the international station was evacuated as a space object approached dangerously toward the station. The length of the space object  is estimated at one-third of an inch, but nevertheless, the body had a great destructive capacity that could threaten the future of the international station estimated at about 100 billion dollars. To solve aircraft-debris collision problem, it is mandatory to move this debris from satellite-orbits toward low orbits close to the earth to burn in the earth’s atmosphere. We can model this problem using Orbital Petri Net Model in Fig.9. Fig.9(a) shows a rotating satellite and a particle in different orbits. The challenge her is how to avoid the collision between the satellite and the particle?, and how to get ride the particle to prevent it making another collisions in the space orbits?.  Fig.9(b) depicts an OPN model for simulating avoiding collision between the satellite $S$ and Debris $D$.  Firing transition $t1$ depends on achieving three conditions , (1)  $M(P1)\geq 1$ and $ M(P1)\equiv w(p1\rightarrow t1)$. (2)  $M(P2)\geq 1$ and $ M(P2)\equiv w(P2\rightarrow t1)$ . (3) the probability of collision is more than 0. After firing $t1$, the token $S$ in $P1$ moves to $P3$, and token $D$ in $P2$ moves to $P4$. In this case $t2$ and $t3$ are enabled, thereby, firing $t3$ require only $M(P4)\geq 1$ and $ M(P4)\equiv w(p4\rightarrow t3)$ which get ride of  the token $D$ and consume it. On the other hand, firing $t2$ require only $M(P3)\geq 1$ and $ M(P3)\equiv w(P3\rightarrow t2)$. Hence, firing $t2$ will return back the token $s$ to $P1$. This scenario simulates returning back the satellite to its valid orbit after avoiding the collision with the particle. Fig. 10(a) depicts firing $t1$, Fig.10$(b)$ depicts firing $t2$, and $t3$.

%=========== Fig 9 ================
  \begin{figure}[H]
\centering
\fcolorbox{black}{white}{\includegraphics[width=1\linewidth]{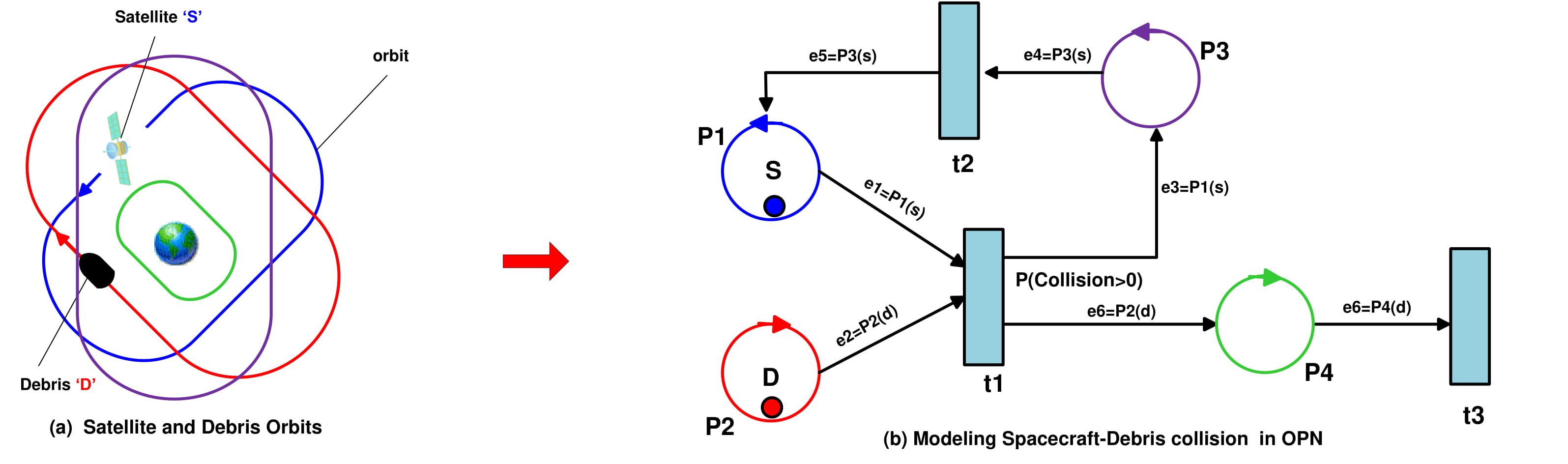}}
\caption{Modeling Spacecraft-Debris collisions in Orbital Petri Net }
\label{fig:1} % Give a unique label
\end{figure}

%==============================

%=========== Fig 10 ================
  \begin{figure}[H]
\centering
\fcolorbox{black}{white}{\includegraphics[width=1\linewidth]{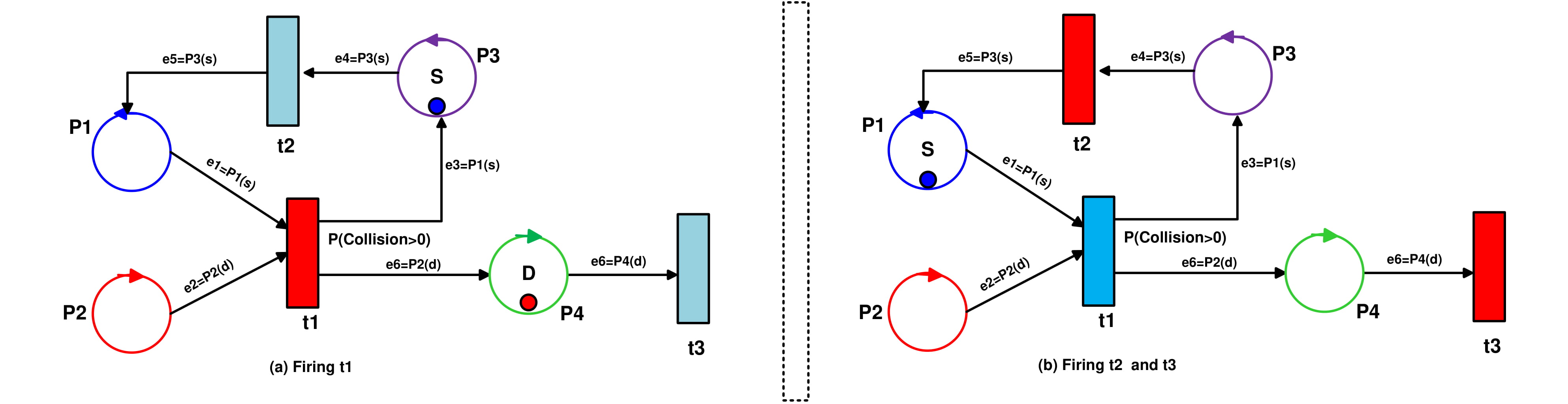}}
\caption{Spacecraft-Debris collision prevention: (a) Firing $t1$ and (b) Firing $t2$ and $t3$ }
\label{fig:1} % Give a unique label
\end{figure}

%==============================

 Mathematically, we can represent the collision prevention between satellite and debris by solving equation 9 , which formulate the Necessary Reachability Condition. Firing transition $t1$, will produce the marking state  $M_1$ , which can be formulated as:

 \begin{equation}\label{EQ 15}
   M_1=\left(
         \begin{array}{c}
           P_1 \\
           P_2 \\
           P_3 \\
           P_4 \\
         \end{array}
       \right)=\left(
                 \begin{array}{c}
                   S \\
                   D \\
                   0 \\
                   0 \\
                 \end{array}
               \right)+\left(
                         \begin{array}{ccc}
                           -S &~~S &~~0 \\
                           -D &~~0 &~~0 \\
                           ~S &~-S &~~0 \\
                           ~D &~~0 &~-D \\
                         \end{array}
                       \right)\times\left(
                                      \begin{array}{c}
                                        1 \\
                                        0 \\
                                        0 \\
                                      \end{array}
                                    \right)
 \end{equation}

 So,

 \begin{equation}\label{EQ 16}
   M_1=\left(
         \begin{array}{c}
           P_1 \\
           P_2 \\
           P_3 \\
           P_4 \\
         \end{array}
       \right)=\left(
                 \begin{array}{c}
                   S \\
                   D \\
                   0 \\
                   0 \\
                 \end{array}
               \right)+\left(
                         \begin{array}{c}
                           -S \\
                           -D \\
                           ~S \\
                           ~D \\
                         \end{array}
                       \right)=\left(
                                      \begin{array}{c}
                                        0 \\
                                        0 \\
                                        S \\
                                        D \\
                                      \end{array}
                                    \right)
 \end{equation}

 Equation 16 represents moving token $S$ from $P1$ to $P3$, and moving token $D$ from $P2$ to $P4$.\\
 Firing transition $t2$, and $t3$ will produce the marking state $ M_2$ , which can be formulated as:

 \begin{equation}\label{EQ 15}
   M_1=\left(
         \begin{array}{c}
           P_1 \\
           P_2 \\
           P_3 \\
           P_4 \\
         \end{array}
       \right)=\left(
                 \begin{array}{c}
                   0 \\
                   0 \\
                   S \\
                   D \\
                 \end{array}
               \right)+\left(
                         \begin{array}{ccc}
                           -S &~~S &~~0 \\
                           -D &~~0 &~~0 \\
                           ~S &~-S &~~0 \\
                           ~D &~~0 &~-D \\
                         \end{array}
                       \right)\times\left(
                                      \begin{array}{c}
                                        0 \\
                                        1 \\
                                        1 \\
                                      \end{array}
                                    \right)
 \end{equation}

 So,
 \begin{equation}\label{EQ 16}
   M_2=\left(
         \begin{array}{c}
           P_1 \\
           P_2 \\
           P_3 \\
           P_4 \\
         \end{array}
       \right)=\left(
                 \begin{array}{c}
                   0 \\
                   0 \\
                   S \\
                   D \\
                 \end{array}
               \right)+\left(
                         \begin{array}{c}
                           S \\
                           0 \\
                           -S \\
                           -D \\
                         \end{array}
                       \right)=\left(
                                      \begin{array}{c}
                                        S \\
                                        0 \\
                                        0 \\
                                        0 \\
                                      \end{array}
                                    \right)
 \end{equation}

 Equation 18 represent returning the token $S$ back to $P1$ and getting ride of the token $D$.

 it is easy to verify and prove the necessary reachability condition in equation
9 as:
 \begin{equation}\label{EQ 16}
   M_2=\left(
                 \begin{array}{c}
                   S \\
                   D \\
                   0 \\
                   0 \\
                 \end{array}
               \right)+\left(
                         \begin{array}{ccc}
                           -S &~~S &~~0 \\
                           -D &~~0 &~~0 \\
                           ~S &~-S &~~0 \\
                           ~D &~~0 &~-D \\
                         \end{array}
                       \right)\times\left(
                                      \begin{array}{c}
                                        1 \\
                                        1 \\
                                        1 \\
                                      \end{array}
                                    \right)
 \end{equation}

 Hence,

\begin{equation}\label{EQ 16}
   M_2=\left(
                 \begin{array}{c}
                   S \\
                   D \\
                   0 \\
                   0 \\
                 \end{array}
               \right)+\left(
                         \begin{array}{c}
                           ~0 \\
                           -D\\
                           ~0 \\
                           ~0 \\
                         \end{array}
                       \right)=\left(
                                      \begin{array}{c}
                                        S \\
                                        0 \\
                                        0 \\
                                        0 \\
                                      \end{array}
                                    \right)
 \end{equation}

 Equation 20 prove that $M_2$ is reachable state from $M_0$. This scenario simulates returning the the satellite back to its valid orbit and consuming the debris for preventing the collision between them.

 \section{Conclusion}
 This paper has introduced a novel approach of Petri Nets called Orbital Petri Net. In this investigation, the aim was to assess the possibility of orbital Petri Nets to study orbital movements of particles within a specific domain. The proposed approach has been analyzed and applied on the space debris collision problem. The mathematical analysis proved that orbital Petri Net can be used to model the solution of this pressing problem. This initial study introduced two models of orbital Petri Nets for simulating preventing two satellite collision problem as well as preventing a satellite and debris collision problem. In the next study, we will investigate in more implementation details new smart algorithms which  can be simulated by orbital Petri Nets for mitigating the space debris collision problem.

\section*{Acknowledgment} This research was supported by an TEDDSAT Project grant, Egypt.

\end{document}